\pdfoutput=1

\documentclass[11pt]{article}

\usepackage{acl}

\usepackage{times}
\usepackage{latexsym}
\usepackage{makecell}
\usepackage[T1]{fontenc}

\usepackage[utf8]{inputenc}

\usepackage{microtype}
\usepackage{breakurl}
\usepackage{times}
\usepackage{latexsym}

\usepackage{microtype}
\usepackage{hyperref}

\usepackage{graphicx}
\usepackage{xspace}
\usepackage{paralist}
\usepackage{booktabs}
\usepackage{multirow}
\usepackage{amsmath}
\usepackage{amsfonts}
\usepackage{amssymb}
\usepackage{pifont}
\usepackage{mathrsfs}
\usepackage{algorithm,algpseudocode}
\usepackage{mdwlist}
\usepackage{enumitem}
\usepackage{color}
\usepackage{dsfont}
\usepackage{amsthm}
\usepackage{bm}
\usepackage{float}
\usepackage{caption}
\usepackage{subcaption}
\usepackage{hyperref}

\usepackage{tikz}
\newcommand*{\circled}[1]{\lower.7ex\hbox{\tikz\draw (0pt, 0pt)%
    circle (.5em) node {\makebox[1em][c]{\small #1}};}}

\newcommand{\expect}{\ensuremath{\mathbb{E}}}

\newcommand{\notes}[1]{}

\theoremstyle{definition}

\theoremstyle{plain}

\newcommand{\vecs}{\ensuremath{\mathbf{s}}\xspace}

\newcommand{\vecu}{\ensuremath{\mathbf{u}}\xspace}

\newcommand{\ith}[1]{\ensuremath{i^{{th}}}}

\newcount\permx
\newcount\permy
\def\permdot#1#2{
\permx=#1 \advance\permx by-1
\permy=#2 \advance\permy by-1
\psframe[fillcolor=black, fillstyle=solid]
(\permx,\permy)(#1, #2)
}

\newcommand\union{\cup}

\newcommand{\ee}{\ensuremath{\mathbf{e}}\xspace}

\newcommand{\boxnum}[1]{{\setlength{\fboxsep}{1pt}\raisebox{1pt}{\hspace{1pt}\fbox{\tiny #1}\hspace{1pt}}}}
\newcommand{\ind}[1]{\ensuremath{_{\kern-0.5pt\boxnum{#1}}}}

\newcommand{\vecy}{\mathbf{y}\xspace}

\newcommand{\smallnt}[1]{\ensuremath{_{\mbox{\tiny PP}}}\xspace}

\newcommand{\pseudocode}{Algorithm}
\floatname{algorithm}{\pseudocode}

\iffalse

\else

\fi

\newcommand{\dsy}{D_{\vecs,\vecy}\xspace}

\newcommand{\duy}{D_{\vecu,\vecy}\xspace}

\newcommand{\dy}{D_{\vecy}\xspace}
\newcommand{\RNum}[1]{\uppercase\expandafter{\romannumeral #1\relax}}

\newcommand{\method}{DUB\xspace}

\begin{document}

\renewcommand{\thefootnote}{\alph{footnote}}

\title{\method: Discrete Unit Back-translation for Speech Translation}

\author{
    Dong Zhang\textsuperscript{\rm 1}\protect\footnotemark[2]~,
    Rong Ye\textsuperscript{\rm 2},
    Tom Ko\textsuperscript{\rm 2},
    Mingxuan Wang\textsuperscript{\rm 2}\protect\footnotemark[1]~,
    Yaqian Zhou\textsuperscript{\rm 1}\protect\footnotemark[1] \\
    \textsuperscript{\rm 1} School of Computer Science, Fudan University~~\\
    \textsuperscript{\rm 2} ByteDance AI Lab \\
    {\tt 	dongzhang22@m.fudan.edu.cn} \\
    {\tt \{yerong, tom.ko, wangmingxuan.89\}@bytedance.com} \\
    {\tt 	zhouyaqian@fudan.edu.cn} \\
}
\maketitle

\renewcommand{\thefootnote}{\fnsymbol{footnote}}
\footnotetext[2]{Work was done while Dong Zhang was a research intern at ByteDance AI Lab.}
\footnotetext[1]{Corresponding author}

\setcounter{footnote}{0}
\renewcommand{\thefootnote}{\arabic{footnote}}

\begin{abstract}
How can speech-to-text translation (ST) perform as well as machine translation (MT)?
The key point is to bridge the modality gap between speech and text so that useful MT techniques can be applied to ST.
Recently, the approach of representing speech with unsupervised discrete units yields a new way to ease the modality problem.
This motivates us to propose \textbf{D}iscrete \textbf{U}nit \textbf{B}ack-translation~(\textbf{\method}) to answer two questions:
\begin{inparaenum}[(1)\upshape]
    \item Is it better to represent speech with discrete units than with continuous features in direct ST?
    \item How much benefit can useful MT techniques bring to ST?
\end{inparaenum}
With \method, the back-translation technique can successfully be applied on direct ST and obtains an average boost of 5.5 BLEU on MuST-C En-De/Fr/Es.
In the low-resource language scenario, our method achieves comparable performance to existing methods that rely on large-scale external data.
Code and models are available at 
\url{https://github.com/0nutation/DUB}.




\end{abstract}

\section{Introduction}

Speech-to-text translation (ST) converts the spoken source language into the written target language, which is a closely related task to machine translation (MT).
In recent years, direct ST that does not rely on intermediate transcription has received considerable attention due to its potential applications in unwritten language scenarios and various domains~\cite{berard2016listen,sung2019towards,han-etal-2021-learning,papi2021speechformer,fang2022stemm,ye2022cross, cheng2023m}. 
One of the major challenges faced by ST is data scarcity, which is similar to the low-resource scenarios encountered in MT. 
Intuitively, techniques developed for low-resource MT~\cite{imamura2018enhancement,xia2019generalized,chen2020facebook,liu2020multilingual,tang2020multilingual} should be utilized to improve ST performance.
However, these techniques are hard to be transferred to ST due to the modality gap between speech and text, where ST takes continuous speech as input and MT takes discrete tokens as input.
Generally speaking, if there is a way to efficiently remove the modality gap, a large number of useful NLP techniques can be applied and facilitate the improvement of ST.


Recently, representing speech with unsupervised discrete units has become popular and successful in the field of speech processing~\cite{baevski2019vq,baevski2020wav2vec,hsu2021hubert,lakhotia2021generative}.
Instead of losing relevant information, discretizing continuous speech has been found to have the advantage of filtering out extraneous signals~\cite{sicherman2023analysing, lakhotia2021generative}, leading to significant improvements in the speech tasks, such as automatic speech recognition~\cite{meng2022cobert}, text-to-speech~\cite{dunbar2019zero}, and speech-to-speech translation~\cite{zhang2021uwspeech,lee2022textless,inaguma2022unity}.
Based on this observation, we are motivated to explore the answers to the following two questions: 
\begin{inparaenum}[(1)\upshape]
    \item Is it better to represent speech with discrete units and use them as model input than with continuous features for direct ST?
    \item By narrowing the modality gap with discrete speech units, how much benefit can useful MT techniques bring to direct ST?
\end{inparaenum}

In this paper, we propose \textbf{D}iscrete \textbf{U}nit \textbf{B}ack-translation~(\textbf{\method}), which migrates the useful back-translation technique from MT to ST by discretizing the speech signals into unit sequences.
In our proposed method, we first convert speech into discrete units using the clustering indices on HuBERT~\cite{hsu2021hubert} representations. 
To complete the translation task, we feed the discrete units into the \textbf{U}nit-\textbf{to}-\textbf{T}ext \textbf{T}ranslation (U2TT) model.
For the back-translation training strategy, \method employs a text-to-unit translation model that learns to predict the source discrete units from the target text.
By leveraging the additional easily accessible text in the target language, we utilize the synthetic parallel data generated by the text-to-unit translation model in conjunction with the original parallel data to update the final unit-to-text model.



Our contributions include the following.
\begin{itemize}[itemsep=1pt, leftmargin=10pt, parsep=0pt, topsep=1pt]
    \item We design a novel unit-text translation (U2TT) framework for direct ST by discretizing the speech feature in an unsupervised manner.
    Our analysis shows that in such a framework, the unit retains the semantic information for translation and can be used as model input.

    \item 
    Based on the U2TT framework, we propose \method, which successfully applies the back-translation technique to direct ST.
    Experimental results show that DUB can further yield an average 5.5 BLEU gain over the U2TT model on the MuST-C English-to-German, French, and Spanish translation directions.

    \item Our approach is particularly beneficial for low-resource or unwritten languages in the world because unit extraction does not require any textual supervision and only speech-translation pairs are used for training.
\end{itemize}

     

\begin{figure*}[t] 
    \setlength{\abovecaptionskip}{-0.cm}
    \setlength{\belowcaptionskip}{-0.5cm}
    \centering 
    \includegraphics[width=0.85\textwidth]{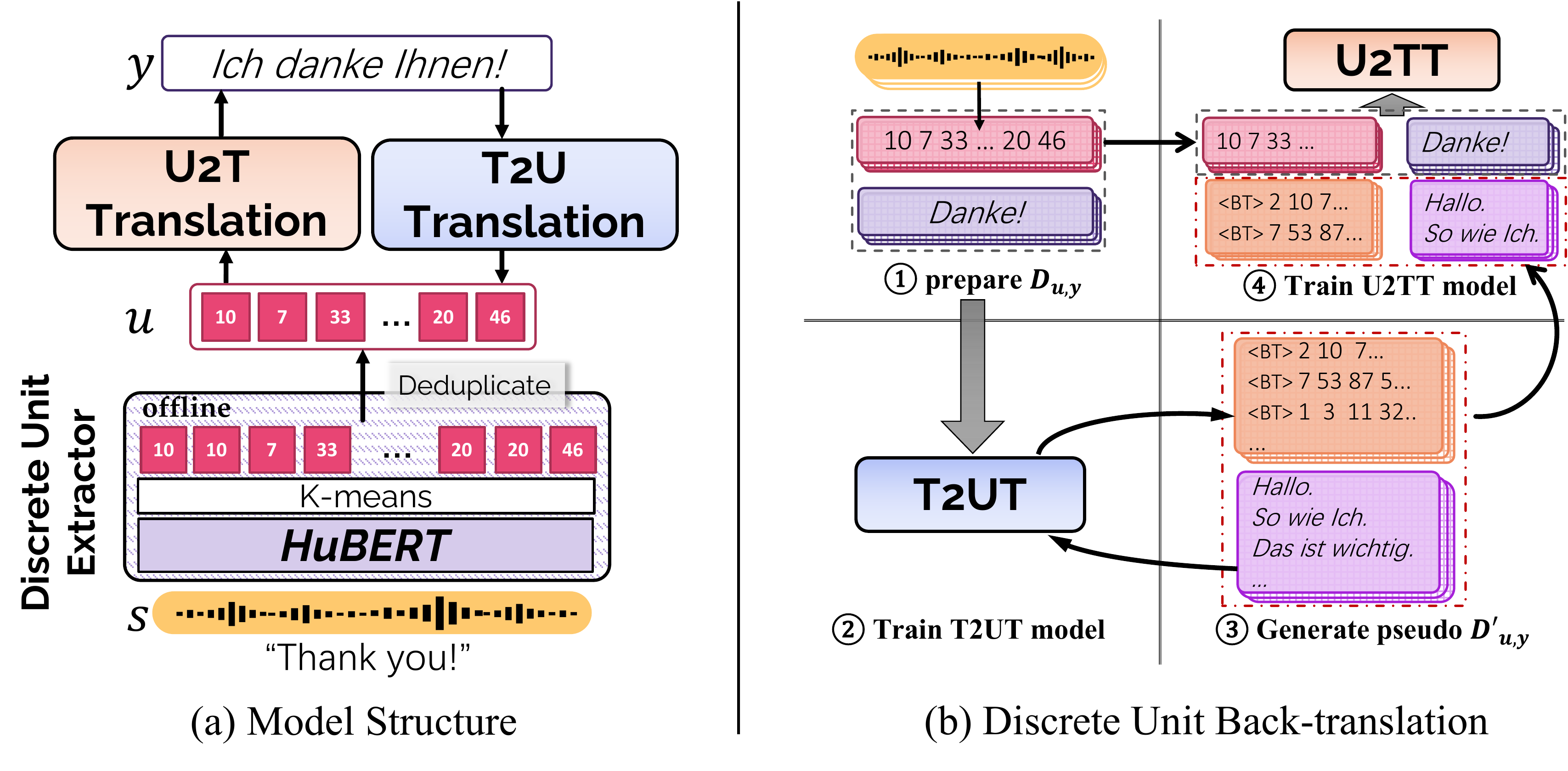} 
    \caption{\textbf{Left}: The model structure of our approach. The offline discrete unit extractor converts speech into discrete units. The unit-to-text translation~(U2TT) model translates the discrete units into the translation and the text-to-unit~(U2TT) model does the opposite. \textbf{Right}: An illustration of the discrete unit back-translation~(\method) training procedure.}
    \label{fig:u2tt_dub} 
\end{figure*}

\section{Related Work}

\noindent\textbf{Speech translation}~
Without using textual transcriptions during inference or training, translating audio directly into the target language is very meaningful for languages that do not have a written form. \citet{berard2016listen} first proposed an end-to-end encoder-decoder architecture for such direct speech-to-text translation. Later, novel models~\cite{di2019adapting,dong2021consecutive,dong2021listen,zheng2021fused} and training techniques, such as multi-task learning~\cite{indurthi2021task,tang2021improving,ye2021end}, knowledge distillation~\cite{liu2019end, dong2021listen}, and pretraining methods~\cite{zheng2021fused, zhang2022speechut}, were developed to improve end-to-end performance. However, these training methods often rely on the use of source text or knowledge from the pre-trained models. 
Without using transcripts or pretraining, \citet{zhang2022revisiting} proposed the parameterized distance penalty to better model speech locality in the self-attention structure and provided results on ST benchmarks covering 23 languages.

\noindent\textbf{Back-translation}~
Back-translation (BT) is a widely used method for improving machine translation systems by training a target-to-source model and creating synthetic parallel data from monolingual target text. This approach has been shown to be effective in both statistical~\cite{bertoldi2009domain,bojar2011improving} and neural machine translation models~\cite{sennrich-etal-2016-improving, edunov-etal-2018-understanding,hoang2018iterative}, and is frequently used to improve translation performance in WMT competitions~\cite{farhad2021findings,wenzek2021findings,adelani2022findings}.
A similar data augmentation idea through synthesizing speech from utterances can be applied to automatic speech recognition (ASR)~\cite{tjandra2017listening, hayashi2018back, ueno2021data}.
However, applying BT in ST is not trivial. \citet{zhang2022ustc} augmented the triplet data by TTS generation from transcription, but the experiment showed that such scaling yields minimal improvement to the final ST model.

\noindent\textbf{Discrete speech representation}~
Discrete speech representations are often studied in the work on self-supervised speech representation learning~\cite{van2017neural,baevski2019vq,baevski2020wav2vec,hsu2021hubert,meng2022cobert}. For example, \citet{van2017neural} proposed Vector Quantised-Variational AutoEncoder (VQ-VAE) to map continuous signals, like speech or image, into a discrete sequence space.
\citet{hsu2021hubert} proposed HuBERT, which learns self-supervised speech representation by extracting speech features and clustering them offline, and iteratively training the clustering indexes of features at masked locations.
Although the clustered discrete representations are only a by-product of HuBERT, they are used to build the generative spoken language
model~\cite{lakhotia2021generative,kharitonov2022text},
enhance speech representation~\cite{chung2021w2v,meng2022cobert,chen2022wavlm,wu2022wav2seq,zhang2022speechut}, 
and model direct speech-to-speech translation~\cite{lee2022textless,inaguma2022unity}.
In the prior literature, probably the most similar task to ours is the textless speech-to-speech translation~\cite{lee2022textless}, but the difference is that they discretized the target-side speech and convert speech-to-speech generation into speech-to-discrete-unit generation, while we discretize the speech at the source side.
\citet{zhang2022speechut} leveraged the discrete unit as an interface to align speech and text, and proposed a unified-modal encoder-decoder pre-training model, SpeechUT.
SpeechUT aims to improve speech representation via the units, while we use the units to construct a unified framework for ST and MT, and to explore transferable training methods from NLP to speech.

\section{Our Approach}

\subsection{Problem Formulation}
\label{sec:031_problem}

Unlike cascade systems or existing end-to-end ST work that utilizes speech-transcription-translation triplet $(\mathbf{s},\mathbf{x},\mathbf{y})$, we aim to build and train the model that translates speech directly into text in another language without using the transcription $\mathbf{x}$. 
The training dataset is denoted as $\mathcal{\dsy} =\{(\mathbf{s}, \mathbf{y})\}$.
Also, we introduce the monolingual corpus of the target language $\mathcal{\dy^{\prime}} = \{\mathbf{y^{\prime}}\}$,  enhance the model via the discrete unit back-translation~(\method) method (described in Section~\ref{sec:033_dub}).


\subsection{Model Structure}
\label{sec:032_model_structure}

As illustrated in Figure~\ref{fig:u2tt_dub}(a), our model consists of three main components: \textit{discrete unit extractor}, \textit{unit-to-text translation model}, and \textit{text-to-unit translation model}. 

\noindent\textbf{Discrete Unit Extractor}~
The discrete unit extractor converts continuous speech signals into a sequence of discrete units, which we use the Hidden-unit BERT~(HuBERT)~\cite{hsu2021hubert}.
HuBERT is a self-supervised model learned by predicting discrete labels of masked audio segments from k-means clustering on the model's intermediate representations. 
It consists of a stack of 1-D convolutional layers and a Transformer encoder to encode the speech into continuous intermediate representations, and a k-means model to convert the representations into a sequence of cluster indices.
We then remove the adjacent duplicate indices to obtain the discrete units sequence, denoted as $\vecu=(\vecu_1, \vecu_2, \ldots, \vecu_T)$, $\vecu_i \in \{0,1, \ldots, K-1\}$, $\forall 1 \leq i \leq T$, where $K$ is the number of clusters.
Note that the discrete unit extractor used \textbf{offline} during the pre-processing stage before translation, can be considered as a \textbf{feature extractor}.

\noindent\textbf{Unit-to-Text Translation (U2TT) Model}~
The U2TT model $\theta_{u \rightarrow y}$ performs the forward translation. It consists of a discrete unit embedding layer and a Transformer. 
The discrete unit embedding layer converts discrete units $\vecu$ into the embedding $\ee=(\ee_1, \ee_2, \ldots, \ee_T)$. 
In order to retain more contextual and textual information from HuBERT, we adopt the intermediate representations of HuBERT's k-means cluster centroids as prior knowledge to initialize the unit embedding. 
This initialization operation is referred to as \textit{pre-trained embedding} in the later analysis~(Section~\ref{sec:unit_infoloss}).
The Transformer follows the vanilla Transformer architecture~\cite{vaswani2017attention}, consisting of a Transformer encoder and a Transformer decoder. The encoder takes unit embedding $\ee$ plus sinusoidal positional embedding as input and outputs semantic representation. The decoder generates the translation sequence $\vecy=(\vecy_1, \vecy_2, \ldots, \vecy_{|\vecy|})$ autoregressively based on the semantic representation.

\noindent\textbf{Text-to-Unit Translation (T2UT) Model}~
The T2UT model $\theta_{y \rightarrow u}$ has the same structure as the U2TT model, but with a randomly initialized text embedding layer. It is added to perform the text-to-unit translation and to incorporate the \method training.

\subsection{Discrete Unit Back-Translation (\method)}
\label{sec:033_dub}

\noindent\textbf{Training Steps}~
Given ST parallel dataset $\mathcal{\dsy} =\{(\mathbf{s}, \vecy)\}$, extra target-language corpus $\mathcal{\dy^{\prime}} = \{\vecy^{\prime}\}$, and the discrete unit extractor $E$. 
As shown in Figure\ref{fig:u2tt_dub}(b), the \method training steps are as follows.

\begin{itemize}[itemsep=1pt, leftmargin=12pt, parsep=0pt, topsep=1pt]
  \item[\textbf{1.}]
  Extract unit for each speech input~$\vecu=E(\mathbf{s})$, and get unit-translation pairs~$\mathcal{\duy}=\{(\vecu, \vecy)\}$;
  
  \item[\textbf{2.}]
  Train the T2UT model based on $\mathcal{\duy}$ with cross-entropy loss as in Eq.~(\ref{eq:2});
  
  \item[\textbf{3.}]
  For each text $\vecy^{\prime} \in \mathcal{\dy^{\prime}}$, generate corresponding synthetic units $\hat{\vecu}^{\prime}$ through the BT model~(generation methods will be discussed in Section~\ref{sec:034_gen_methods}).
  Then, add special \texttt{<BT>} indicator at the begining of $\hat{\vecu}^{\prime}$~\cite{caswell2019tagged}. 
  The synthetic units-translation set is denoted as $\mathcal{\duy^{\prime}} =\{(\hat{\vecu}^{\prime}, \vecy^{\prime})\}$;
  
  \item[\textbf{4.}]
  Upsample the original training data by a rate of $r$ and train U2TT model based on $\mathcal{\duy}\union \mathcal{\duy^{\prime}}$ and loss in Eq.~(\ref{eq:1}).
\end{itemize}

\noindent\textbf{Training Objective}~
The training objectives for U2TT and T2UT models are the negative log-likelihood losses based on the unit-translation pairs:
\begin{align}
     \mathcal{L}_{\text{U2TT}} = -\expect_{(u,y) \in D} \log P\left(\vecy \mid \vecu,\theta_{u\to y}\right)
     \label{eq:1} \\
     \mathcal{L}_{\text{T2UT}} = -\expect_{(u,y) \in D_{u,y}} \log P\left(\vecu \mid \vecy,\theta_{y\to u}\right)
     \label{eq:2}
\end{align}
, where 
$D$ refers to $\mathcal{\duy} \union \mathcal{\duy^{\prime}}$ for \method training, and when $D=\mathcal{\duy}$, Eq.~(\ref{eq:1}) is the loss function for training U2TT from scratch.

\subsection{Generation methods for back-translated units}
\label{sec:034_gen_methods}
We explore the following generation methods for producing synthetic units: beam search, sampling, and top-k sampling.  We also apply a speech normalization method to remove speaker information when generating units.

\textbf{Beam search} tries to identify the maximum a posteriori~(MAP) output and generate the sentence with the largest estimated probability given an input. 
\textbf{Sampling} means sampling from the distribution randomly at each step, which generates diverse outputs. 
\textbf{Top-k sampling} is a middle ground between beam search and sampling. At each time step, we select the \emph{k} most likely tokens from the output distribution, re-normalize and then sample from this restricted set.

The discrete unit extractor produces various unit sequences for speech with the same content when delivered by multiple speakers~\cite{lee2022textless}.  These variations pose a challenge for training the BT model. In order to address this issue, we adopt a \textbf{Speech Normalization} module from~\cite{lee2022textless}, which removes speaker information from the discrete units and produces norm units. 
Specifically, it is an off-the-shelf HuBERT-CTC model trained on VoxPopuli~\cite{wang2021voxpopuli} that normalizes variant speech input to a single speaker to eliminate such influence (denoted as \textit{Speech Norm}). We implement back-translation with norm units and use the resulting BT model to generate pseudo norm units.

\section{Experiments}

\subsection{Datasets}
\label{sec:041_datasets}

\noindent\textbf{MuST-C}~
MuST-C\footnote{All released under CC BY NC ND 4.0 International\label{fn:data}}~\cite{di2019must}, one of the most widely-used ST benchmarks, contains translations from English to 8 languages collected from TED talks.
We train and validate our approach in three ST directions: English-German~(En-De), English-French~(En-Fr), and English-Spanish~(En-Es).

\noindent\textbf{CoVoST-2 X-En}~
CoVoST-2~\cite{wang2021covost} is a multilingual ST corpus derived from the Common Voice project, which offers translations from English to 15 languages and from 21 languages to English.
We conducted experiments on X-En, including high-resource languages~($\geq$ 10 hours of speech) such as French~(Fr) and German~(De), and low-resource languages~($\leq$ 2 hours of speech), like Arabic~(Ar), Swedish~(Sv), Japanese~(Ja), etc.
Without the need for transcription, the evaluation focuses on the capability of our method to generalize to the low-resource unwritten multi-languages. 

\noindent\textbf{Monolingual text corpus}~
Monolingual target-language text corpora are introduced for back-translation. 
For MuST-C we include 48M German, 79M French and 64M Spanish sentences sampled from TED~\cite{duh18multitarget}, WMT~\cite{bojar-EtAl:2016:WMT1} and CCMartix~\cite{schwenk2019ccmatrix} datasets for En-De/Fr/Es respectively.
For CoVoST-2 X-En experiments, we introduce 1M extra English sentences sampled from the transcriptions of Common Voice Corpus 11.0~\cite{ardila-etal-2020-common}.

All statistics of the datasets are in Appendix~\ref{sec:app:data statistics}.

\subsection{Experimental setups}
\noindent\textbf{Pre-processing}~
The model accepts 16-bit 16KHz mono-channel raw waveform speech and then discretizes them into units.
We denote the discrete units clusters by the numbers (\textit{e.g.} $\#1$, $\#2$), and combining with the target-language sentences, we learn the joint vocabulary via SentencePiece~\cite{kudo-richardson-2018-sentencepiece}. We set the joint vocabulary size to 8000 for both MuST-C and CoVoST-2. 

\noindent\textbf{Model Configuration}~
For MuST-C experiments, we use the HuBERT-base\footnote{\url{https://dl.fbaipublicfiles.com/hubert/hubert\_base\_ls960.pt}} (pre-trained on Librispeech without fine-tuning) with a 500-cluster k-means quantizer based on the 9th layer representation as the \textit{discrete unit extractor}.
For CoVoST-2, we employ the mHuBERT\footnote{\url{https://dl.fbaipublicfiles.com/hubert/mhubert_base_vp_en_es_fr_it3.pt}} with a 1000-cluster k-means quantizer based on the 11th layer representations pre-trained on the VoxPopuli~\cite{wang2021voxpopuli} speech in English, Spanish, and French.
The U2TT and T2UT models have the same model architecture, consisting of a 12-layer Transformer encoder and a 6-layer Transformer decoder, with hidden size $d=768$, $16$ attention heads, and $4096$ FFN hidden states. 
Additionally, we implement the \textsc{Base} and \textsc{Large} versions of the model, with hidden sizes of 768 and 1024, respectively. Both versions are performed in the main experiments, while in the analysis, we primarily investigate the \textsc{Base} model.
More information on model size and scalability experiments can be found in Appendix~\ref{sec:app:ana_model_data_size}.



\noindent\textbf{Evaluation}~
We evaluate the models using case-sensitive sacreBLEU\footnote{\url{https://github.com/mjpost/sacrebleu}, \textbf{BLEU Signature}:
nrefs:1 | bs:1000 | seed:12345 | case:mixed | eff:no | tok:13a | smooth:exp | version:2.0.0}~\cite{post2018call} on the MuST-C \texttt{tst-COM} sets and CoVoST-2 test sets.

See Appendix~\ref{sec:app:experimental_deatils} for more details on vocabulary learning, training, and test.

\subsection{Baseline models}
\label{sec:043_baselines}
We compare our method with the baselines as listed in Table~\ref{tab:mustc} $\sim$ \ref{tab:covost_low}~(Appendix~\ref{sec:app:baseline_models} for details). In particular, we explain the following baselines that do not involve transcriptions during training.

\noindent\textbf{Revisit ST}~\cite{zhang2022revisiting}
is a direct speech-to-translation model with parameterized distance penalty (PDP) and CTC regularization. Its framework and training objectives are sorely different from ours.

\noindent\textbf{Unit-to-text Translation~(U2TT)}~
has the structure as described in Section~\ref{sec:032_model_structure} and is trained using only speech-translation supervision from the ST dataset from scratch, without applying \method. As a baseline, comparison with this model helps to see the influence of~\method.

\noindent\textbf{Transformer-ST}~
stands for training the Speech-Transformer~\cite{dong2018speech} from scratch, but without ASR pre-training as in the previous work~\cite{wang-etal-2020-fairseq, inaguma2020espnet, zhao2021neurst}. The training details are in Appendix~\ref{sec:app:baseline_models}.





\begin{table}[t]
    \setlength{\belowcaptionskip}{-0.5cm}
    \centering
    \resizebox{1.05\linewidth}{!}{
    \begin{tabular}{l|ccc|c}
        \toprule
        \textbf{Method} & \textbf{De} & \textbf{Fr} & \textbf{Es} & \textbf{Avg}. \\
        \midrule
        \multicolumn{5}{l}{\textit{Methods that utilize transcriptions}} \\
        Fairseq ST~\cite{wang-etal-2020-fairseq} &  22.7 &  32.9 &  27.2 & 27.6 \\
        NeurST~\cite{zhao2021neurst} & 22.8 & 33.3 & 27.4 & 27.8 \\
        Espnet ST~\cite{inaguma2020espnet} & 22.9 & 32.8 & 28.0 & 27.9\\
        E2E-ST-JT~\cite{du2022regularizing} & 23.1 & 32.8 & 27.5 & 27.8 \\
        Speechformer~\cite{papi2021speechformer} & 23.6 & - & 28.5 & - \\

        Cascaded~\cite{inaguma2020espnet} & 23.6 & 33.8 & 28.7 & 28.7\\
        MTL~\cite{tang2021improving} & 23.9 & 33.1 & 28.6 & 28.5 \\
        Self-training~\cite{pino2020selfa} & 25.2 & 34.5 & - & - \\
        SpeechT5~\cite{ao2022speecht5} & 25.2 & \textbf{35.3} & - & - \\
        \midrule
        \multicolumn{5}{l}{\textit{Methods that do not involve transcriptions}} \\
        Revisit ST~\cite{zhang2022revisiting} & 23.0 & 33.5 & 28.0 & 28.2\\
        Transformer-ST &  18.0 & 28.5 & 24.1 & 23.5 \\
        U2TT~(\textsc{Base}) & 20.4 & 30.3 & 25.3 & 25.3 \\
            \quad\quad w/ \method & 25.8 & 34.7 & 30.2 & 30.2 \\
        U2TT~(\textsc{Large}) & 20.5 & 30.1 & 24.7 & 25.1\\
        \quad\quad w/ \method & \textbf{26.2} &  \textbf{35.3} & \textbf{30.4} & \textbf{30.6} \\
        \midrule
        \multicolumn{5}{l}{\textit{SoTA: use much more speech and various pre-training tasks}} \\
        SpeechUT~\cite{zhang2022speechut}$^{\ast}$ & 30.1 & 41.4 & 33.6 & 35.0 \\ 
        \bottomrule
        
    \end{tabular}
    }
    \caption{BLEU Scores on MuST-C En-X \texttt{tst-COM} set. $\ast$ is the state-of-the-art system, which designed various mask-predict pre-training tasks and trained using extra 1.4k hours of speech and parallel MT data from WMT. Random sampling is the decoding strategy for DUB.}
    \label{tab:mustc}
\end{table}

\begin{table*}[]
    \small
    \centering
    \begin{tabular}{l|cc|ccccccc|c}
        \toprule
        & \multicolumn{2}{c|}{\textbf{Aux. Data}} & \\
        \textbf{Methods} & \textbf{ASR} & \textbf{Text} & \textbf{Fr} & \textbf{De} & \textbf{Es} & \textbf{Ca} & \textbf{It} & \textbf{Ru} & \textbf{Zh} & \textbf{Avg.} \\
        \midrule 
        Transformer-ST$^{\dagger}$ & - & - & 
        4.3 & 8.4 & 12.0 & 14.4 & 0.2 & 1.2 & 1.4 & 8.8 \\
        Transformer-ST + ASR pre-train$^{\dagger}$ & \checkmark & - &
        26.3 & 17.1 & 23.0 & 18.8 & 11.3 & 14.8 & 5.8 & 16.7 \\
        
        Cascaded ST$^{\dagger}$ & \checkmark & - & 
        27.6 & \textbf{21.0} & 27.4 & 21.3 & 13.5 & 16.8 & \textbf{7.0} & 19.2\\
        
        Revisit ST~\cite{zhang2022revisiting} & - & - & 
        26.9 & 14.1 & 15.7 & 17.2 & 2.4 & 3.6 & 2.0 & 11.7 \\
        \midrule
        U2TT~(\textsc{Large}) & - & - &
        27.4 & 16.7 & 28.1 & 23.1 & 20.0 & 21.9 & 5.9 & 20.5 \\
        \quad w/ \method & - & \checkmark1M &
        \textbf{29.5} & 19.5 & \textbf{30.9} & \textbf{25.2} & \textbf{23.9} & \textbf{23.2} & 6.1 & \textbf{22.6} \\
        \bottomrule
    \end{tabular}
    \caption{Test BLEU scores on CoVoST-2 X-En language pairs with more than 10 hours of speech.
    Auxiliary data refers to all data at training excluding \textit{<speech,translation>} pairs.
    $^{\dagger}$: Results from \cite{wang2021covost}. Random sampling is the decoding strategy for DUB.
    }
    \label{tab:covost_high}
\end{table*}

\begin{table*}[]
    \setlength{\belowcaptionskip}{-0.4cm}
    \small
    \centering
    \begin{tabular}{l|cc|cccccccc|c}
        \toprule
        & \multicolumn{2}{c}{\textbf{Aux. Data}}& \textbf{Ar} & \textbf{Sv} & \textbf{Lv} & \textbf{Sl} & \textbf{Ta} & \textbf{Ja} & \textbf{Id} & \textbf{Cy} & \textbf{Avg.} \\
        \textbf{Methods} & \textbf{ASR} & \textbf{Text} & 
        2h & 2h & 2h & 2h & 2h & 1h & 1h & 2h & - \\
        \midrule
        Transformer-ST$^{\dagger}$ &  - & - &
        0.3 & 0.2 & 0.1 & 0.3 & 0.3 & 0.3 & 0.4 & 0.3 & 0.3 \\
        Transformer-ST A2E$^{\dagger}$ &  - & - &
        0.6 & 0.6 & 0.4 & 1.2 & 0.1 & 0.2 & 0.3 & 2.6 & 0.8 \\
        Transformer-ST + ASR pre-train$^{\dagger}$ & \checkmark & - &
        4.3 & 2.7 & 2.5 & 3.0 & 0.3 & 1.5 & 2.5 & 2.7 & 2.4 \\
        \midrule
        \multicolumn{12}{l}{
        \textit{Larger models based on large-scale multilingual speech, text or joint pre-training, involving more data}} \\
        XLS-R (0.4B)$^{\ast}$ & \checkmark & - & 
        8.1 & 5.3 & 3.1 & 5.3 & 0.0 & 2.0	& 3.3 & 3.4 & 3.8 \\
        Wav2seq (0.4B)$^{\ast}$ & \checkmark & - &
        \textbf{10.5} & 8.8 & 4.8 & 5.9 & 0.0 & 1.9 & 5.0 & \textbf{5.7} & 5.3 \\
        XLS-R + mBART-50 (0.7B)$^{\diamondsuit}$ & \checkmark & \checkmark & 
        3.0 & \textbf{10.3} & 6.0 & 6.6 & 0.2 & 0.6 & 1.4 & 2.5 & 3.8 \\
        LNA-E,D (0.7B)$^{\spadesuit}$ & \checkmark & \checkmark & 
        3.7 & 5.9 & 4.6 & 4.6 & \textbf{0.7} & 1.7 & 2.9 & 2.8 & 3.4 \\
        \midrule
        U2TT~(\textsc{Large}) & - & - &
        7.0 & 8.0 & 6.3 & 6.8 & 0.3 & 1.6 & 6.6 & 2.7 & 4.9 \\
        \quad w/ \method & - & \checkmark 1M & 
        7.1 & 8.9 & \textbf{6.9} & \textbf{7.9} & 0.5 & \textbf{2.1} & \textbf{7.0} & \textbf{5.7} & \textbf{5.8} \\
        \bottomrule
    \end{tabular}
    \caption{Test BLEU scores on CoVoST-2 low-resource X-En language pairs with less than 2 hours of speech. $^{\dagger}$: results from~\cite{wang2021covost}.
    $^{\ast}$:
    results from~\cite{wu2022wav2seq}
    $^\diamondsuit$: results from~\cite{babu2021xls}. $^\spadesuit$: results from~\cite{li2021multilingual}. The numbers in parentheses are their parameter sizes.
    Random sampling is the decoding strategy for DUB.
    }
    \label{tab:covost_low}
\end{table*}

\subsection{Main results on Speech-to-text Translation}
\noindent\textbf{MuST-C}~
As shown in Table~\ref{tab:mustc}, compared to the methods that do not involve the transcribed text, our method, U2TT~(\textsc{Large}) with \method, gets the best ST results by introducing extra target-language text, and \method obtains an average boost of 5.5 BLEU compared with U2TT in the three En-X directions.
Encouragingly, we find that our method achieves comparable performance to previous models that utilize transcriptions through multi-task learning or pre-training.
As for the baseline, U2TT outperforms the Transformer-ST, where we believe that the discrete units still retain the semantic information of the audio feature (\textit{e.g.} log Mel-filter bank, \textit{abbr.} Fbank) for translation. 
As for the gap between our method and the SoTA system, we argue that SpeechUT~\cite{zhang2022speechut} performed various mask-predict pre-training tasks using extra 1.4k hours of speech and parallel MT data from WMT, which is not included in our approach.

\noindent\textbf{CoVoST-2}~
Our method performs similarly to MuST-C on the high-resource En-X~(Table~\ref{tab:covost_high}).
Without considering auxiliary data or pre-training methods, adding only 1M additional English text, \method improves by an average of 2.1 BLEU over 7 language pairs compared to U2TT, and by an average of 3.4 BLEU over the cascaded ST system.
For the low-resource setting, our method can bring improvement on almost every language pair and achieve better performance than the large-scale multilingual speech or text pre-training models, like XLS-R+mBART-50 model~\cite{babu2021xls}, with much fewer parameters. 
The discrete unit extractor is unsupervised, so our method does not require transcriptions, which is particularly advantageous for unwritten ST. 
This experiment mimics such low-resource nature of unwritten languages in practice. The results also show that the U2TT model and the DUB training have the potential to translate low-resource unwritten languages.


\section{Analysis on the Effect of Discrete Unit Back-translation (\method)}
\subsection{Is DUB better than other methods that leverage extra raw data? }
\label{sec:utilize_raw_text}
\begin{table}[]
    \setlength{\belowcaptionskip}{-0.5cm}
    \centering
    \small
    \resizebox{\linewidth}{!}{
    \begin{tabular}
    {l|cc|ccc}
        \toprule & \multicolumn{2}{c|}
        {\textbf{Aux. Data}} & \\
        \textbf{Method} & 
        \textbf{Speech} & 
        \textbf{Text} & 
        \textbf{BLEU} \\
        \midrule
        Transformer-ST & - & - & 18.0 \\
        \quad w/ Cascaded BT & - & 10M & 20.3 \\
        \midrule 
        U2TT & - & - & 20.4 \\
        \quad w/ Bimodal BART & 10k\textbf{h} & 10M & 22.4 \\
        \quad w/ \method & - & 10M & 25.0 \\
        \quad w/ Bimodal BART + \method & 10k\textbf{h} & 10M & 25.4 \\
        \bottomrule
        
    \end{tabular}
    }
    \caption{MuST-C En-De \texttt{tst-COM} BLEU scores for different methods that utilize 10M monolingual text data in ST. 
    Transformer-ST and U2TT are described in Section~\ref{sec:043_baselines}.
    }
    \label{tab:utilize_raw_text}
\end{table}

The key benefit of the \method is to make use of a lot of monolingual text.
Here, alternative techniques such as pseudo-labeling and pre-training (implemented as Cascaded BT and Bi-modal BART) 
are also evaluated on the MuST En-De translation, by introducing an equivalent corpus of 10 million German sentences.
\begin{itemize}[itemsep=1pt, leftmargin=10pt, parsep=0pt, topsep=1pt]
    \item 
    \textbf{Cascaded BT}~ 
    aims to build a target-to-source MT-TTS pipeline to construct pseudo-speech translation augmented data for the training.
    Specifically, we use transcription-translation pairs of MuST-C to train a back-translation MT model and use the released FastSpeech2\footnote{\url{https://github.com/ming024/FastSpeech2}}~\cite{ren2020fastspeech} and a HiFi-GAN~\cite{kong2020hifi} vocoder for TTS generation. 
    \item 
    \textbf{Bi-modal BART}~
    has the same structure as U2TT, and is pre-trained by denoising large-scale corrupted discrete units and monolingual text, following the recipe of mBART~\cite{liu2020multilingual}.
    We combine the 10M additional text with 7M discrete units extracted from 10k hours of speech in GigaSpeech~\cite{chen2021gigaspeech} to pre-train the model and fine-tune it based on MuST-C unit-translation pairs. 
    See Appendix~\ref{sec:app:experimental_deatils} for training details.
\end{itemize}

As shown in Table~\ref{tab:utilize_raw_text} introducing equivalent raw text, \textbf{DUB is superior to the above two approaches and has a greater potential to exploit monolingual raw text}. 
We find that the gain from \textit{cascaded BT}-synthesized speech is limited because the synthetic speech is robotic and monotonic, making it easy to overfit the model to the synthetic pairs. 
Although the bi-modal BART pre-training can bring about 2 BLEU improvements, it is still inferior to DUB. 
We attribute this to the gap between the denoising pre-training task and the downstream generation tasks, while DUB does not have such a gap.
Meanwhile, we observe that combination of bi-modal BART and DUB can bring further performance improvements, which indicates that they are complementary to each other.

\subsection{The better the pseudo-unit, the more effective the DUB method?}
\label{sec:pseudo_unit_quality}

In Section~\ref{sec:033_dub}, we presented four generation methods to create synthetic pseudo-units based on the BT model, namely beam search, sampling, top-k sampling, and speech normalization. 
In the experiments, we set a beam size of 5 for the beam search, k=10 for top-k sampling, and use an off-the-shelf speech normalizer\footnote{\url{https://dl.fbaipublicfiles.com/fairseq/speech_to_speech/speech_normalizer/en/en_10h.tar.gz}} from \citet{lee2022textless} for Speech Norm. 

\begin{figure}[htb]
    \centering
    \includegraphics[width=0.87\linewidth]{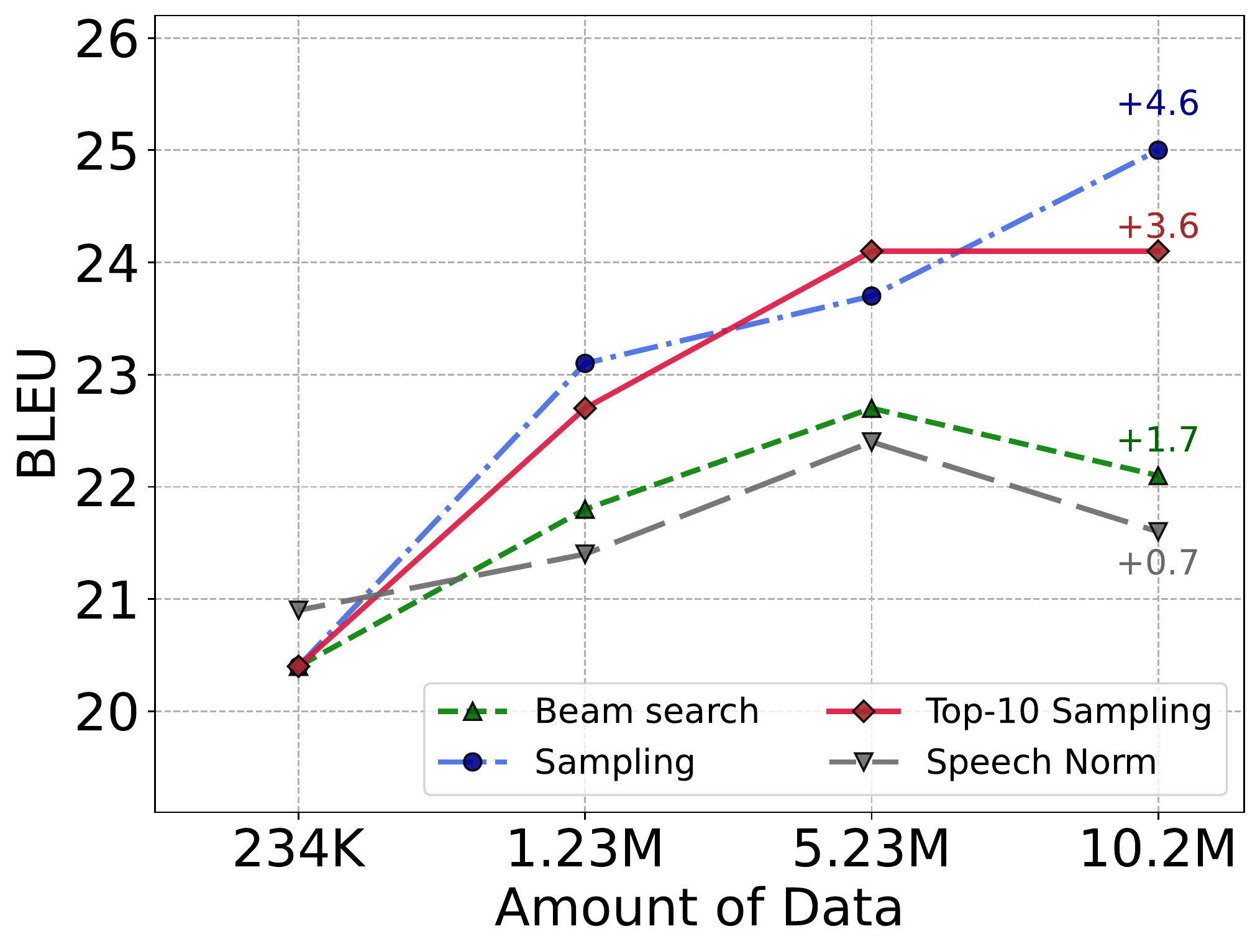}
    \caption{MuST-C En-De \texttt{tst-COM} BLEU scores for the increasing amounts of BT data (1M, 5M, and 10M), obtained using different generation methods. Note that the unit extractor of \textit{Speech Norm} is mHuBERT, so without extra text, the BLEU score is 20.9, which is slightly higher than HuBERT, but the gap is quite small.
    }
    \label{fig:bt_strategy}
\end{figure}

Does the forward model gain more from synthesized pairs when the synthesized units are of higher quality?
We calculate the Unit Error Rate (UER) on the MuST-C validation set to assess the synthesis quality.   
A lower UER indicates that the generated units are closer to the directly extracted units, \textit{i.e.} of higher quality. 
We systematically vary the back-translated data from 1M to 10M, and present the BLEU scores and UERs of the generation methods in Table~\ref{tab:bt_uer} and Figure~\ref{fig:bt_strategy}. 
The \textit{Speech Norm} module produces the highest quality synthesized units, while the sampling-based methods have lower quality. Interestingly, the sampling method with the lowest synthesis quality has the most significant improvement over the forward model.

We conjecture that \textbf{the richness and irregularity of the synthesized data can better improve the forward ST model}, while regular pseudo-units, \textit{e.g.} generated by MAP-based beam search, are more predictable and not conducive to performance improvement. 
This is consistent with previous findings of BT techniques in machine translation~\cite{edunov-etal-2018-understanding}.
In addition, \textbf{\textit{Speech Norm}, which normalizes speech to a single speaker, is not necessary for our \method method}. 
Although such an operation makes the ST model easier to learn and the UER smaller, it compromises the diversity of the synthetic data, which is also not helpful for performance improvement. The model generalization ability weakens when these single-speaker synthesis units increase. 

\begin{table}[htb]
    \setlength{\belowcaptionskip}{-0.2cm}
    \centering
    \begin{tabular}{lrr}
        \toprule
         & \textbf{UER(\%)} & \textbf{$\Delta$BLEU}\\
        \midrule
        Speech Norm & 73.0 & 0.7 \\
        Beam Search & 83.0 & 1.7 \\
        Top-10 Sampling & 89.0 & 3.6\\
        Sampling & 92.0 & 4.6 \\
        \bottomrule
    \end{tabular}
    \caption{The quality of generated pseudo-units using different generation methods and their BLEU increases from 10M extra texts,  evaluated by \textbf{Unit Error Rate (UER)} on MuST-C En-De \texttt{Dev}, the smaller the better. 
    }
    \label{tab:bt_uer}
\end{table}

\section{Why does \method work? --- Analysis on the Property of Discrete Unit}
Why does DUB work? To answer this question, we examine the properties of the discretized speech unit. Specifically, (1) do the units make sense to replace the original speech input in the forward translation process? (2) Do the units generated by back translation also contain semantic information and can they even restore the speech?

\subsection{Are discrete units suitable features for ST input?}
\label{sec:unit_infoloss}

We show how much semantic information is retained for different input forms by comparing the results of the downstream ST task~(shown in Table~\ref{tab:discrete_infoloss}).
Training from scratch, we find that the U2TT model translates better than Transformer-ST~(19.9 \textit{vs.}18.0), 
indicating that \textbf{compared to the speech feature, like Fbank, the discrete unit is a better choice for input and has no information loss in terms of the semantic information required for translation.}
We assume that this is strongly correlated with the HuBERT-based discrete unit extractor, since HuBERT is designed to learn a combined acoustic and language model over the continuous speech input, which preserves much textual information for the speech.
But rigorously, compared to the continuous representation of HuBERT, the discretization procedure does suffer from semantic information loss. Comparing Line~\RNum{2} and \RNum{3}, there is a gap of 2.9 BLEU between U2TT and HuBERT-Transformer (where frozen HuBERT Layer-9 continuous representation is taken as input to perform ST), in terms of ST metrics. 
Fortunately, the gap can be compensated by 
\begin{inparaenum}[\itshape a)\upshape]
    \item initializing the unit embedding as its corresponding K-means cluster centroid on continuous HuBERT representations as described in Section~\ref{sec:032_model_structure}~(denoted as pre-trained embedding, Line~\RNum{4}), which can slightly close the gap by 0.5 BLEU;
    and
    \item simply introducing only 1M additional text and applying \method, which can achieve 2.7 BLEU improvement (Line~\RNum{5} \textit{vs.} \RNum{4}).
\end{inparaenum}

\begin{table}[]
    \setlength{\belowcaptionskip}{-0.5cm}
    \centering
    \begin{tabular}{cl|l}
        \toprule
        \textbf{No.} & \textbf{Methods} &   \textbf{BLEU} \\
        \midrule
        \RNum{1} & Transformer-ST & 18.0 \\
        \RNum{2} & HuBERT + Transformer  & 22.8 \\
        \midrule
        \RNum{3} & U2TT & 19.9 \\
        \RNum{4} & $\hookrightarrow$ w/ pre-trained embedding & 20.4$^{\ast}$ \\
        \RNum{5} & \qquad $\hookrightarrow$ w/ \method-1M & 23.1 \\
        
        \bottomrule
    \end{tabular}
    \caption{BLEU scores on MuST-C En-De \texttt{tst-COM} set. 
    DUB-1M means to introduce 1M unpaired text via the DUB method.
    $^\ast$:~the improvement over U2TT is statistically significant~($p<0.05$)}
    \label{tab:discrete_infoloss}
\end{table}


\subsection{Can we recover faithful speech from pseudo-units?}
\label{sec:text-to-speech translation}

Do the back-translated units capture the semantics of the target language text? 
Since it is difficult to directly evaluate the correctness of the pseudo-units generated by the back-translation model, we concatenate a unit-based HiFi-GAN vocoder\footnote{\url{https://dl.fbaipublicfiles.com/fairseq/speech_to_speech/vocoder/code_hifigan/mhubert_vp_en_es_fr_it3_400k_layer11_km1000_lj/g_00500000}} with our back-translation model to recover speech from the generated pseudo-units, thus completing the text-to-speech~(TTS) translation task.
TTS generation quality is measured by ASR-BLEURT, where we transcribe the speech output using a high-quality open-source ASR model\footnote{\url{https://huggingface.co/facebook/wav2vec2-large-960h-lv60-self}} and calculate BLEURT\footnote{\url{https://github.com/google-research/bleurt}} with reference transcription.
As shown in Table~\ref{tab:tts_translation}, the ASR-BLEURT of beam search and sampling is 0.6 and 0.47 respectively, indicating that \textbf{the unit sequence back-translated from a given target language text can convey its general semantic meaning, which is the guarantee for the success of \method.}
We conduct the listening test by checking 30 randomly sampled BT-recovered speeches for semantic consistency with ground-truth. 22 of 30 sentences matched ground-truth speech, while the remaining 8 had minor issues, with only 1 being of low quality and the other 7 missing or repeating 1-2 details.
We also provide some generated audio samples in Appendix~\ref{sec:app:cases_tts_translation} to help illustrate the degree of speech restoration.

\begin{table}[htb]
    \setlength{\belowcaptionskip}{-0.5cm}
    \centering
    \begin{tabular}{l|ccc}
        \toprule
        \textbf{Generation Method} & \textbf{ASR-BLEURT} \\
        \midrule
        beam search & 0.60 \\
        sampling & 0.47 \\
        \bottomrule
        
    \end{tabular}
    \caption{MuST-C En-De \texttt{tst-COM} ASR-BLEURT for text-to-speech translation}
    \label{tab:tts_translation}
\end{table}


\section{Conclusion}
In this paper, we propose \textbf{D}iscrete \textbf{U}nit \textbf{B}ack-translation (DUB), as well as the \textbf{U}nit-\textbf{to}-\textbf{T}ext \textbf{T}ranslation (U2TT) model for direct speech translation.
Our approach successfully migrates the back-translation technique from MT to ST by discretizing the speech signals into unit sequences and making use of extra widely accessible text in the target language. 
Without using transcription, \method can achieve an average increase of 5.5 BLEU on MuSTC En-De/Fr/Es over the raw U2TT framework, and achieves comparable performance to the large-scale speech-text joint pre-training models on CoVoST-2 low-resource ST.
The analysis experiments also show the potential of such discrete audio units as inputs and outputs for text or speech generation tasks.

\section*{Broader Impact}
Our proposed model structure with a discrete unit extractor for speech and the unit-to-text translation model, which does not need any transcriptions during training, is particularly relevant for speech translation for more than 3,000 languages and dialects in the world that cannot be transcribed.
Since these unwritten languages are typically low-resource, we emphasize that boosting ST performance via text-to-unit back-translation data augmentation, \textit{i.e.} \method, is very promising.
Meanwhile, as a by-product of \method, TTS translation has significant implications for assisting visually impaired or dyslexic people in understanding the world as well as preserving low-resource unwritten spoken languages.

However, as exploratory work, we focus on investigating the potential of using BT to enhance ST performance, while popular large-scale pre-training methods are not employed in this paper. 
This makes our method slightly inferior to these methods in terms of performance, perhaps.
But promisingly, in terms of structure, the model is more general across various modalities and also has more potential to integrate with the methods in NLP area (might be the topic of future research). 
Also, the models are still far from real industrial applications. For example, the data used for training is much smaller than the scale in reality, while the real speech is noisier and more complex than the open-source dataset, which may require front-end processing.
Moreover, the success of our method is partly attributable to the HuBERT representation, which contains certain textual information for the speech, and via experiments, we also find that the quality of discrete units influences the translation performance. Nevertheless, learning meaningful discrete units is not the primary goal of HuBERT pre-training, and how to learn discrete units or representations for speech with more \textit{contextual} semantic information can be explored in the future. 

\section*{Acknowledgements}
We thank Fuliang Weng for the careful guidance and revisions to the paper and thank all the anonymous reviewers for their insightful and valuable comments.


\bibliography{custom}
\bibliographystyle{acl_natbib}

\clearpage
\appendix

\section{Data Statistics}
\label{sec:app:data statistics}

\begin{table}[h]
    \centering
    \begin{tabular}{c|cc}
    \toprule
    \textbf{En$\rightarrow$} & \textbf{Hours} & \textbf{Samples} \\
    \midrule
    \textbf{De} & 408 & 234K \\
    \textbf{Fr} & 492 & 280K \\
    \textbf{Es} & 504 & 270K \\
    \bottomrule
    \end{tabular}
    \caption{Statistics of MuST-C dataset}
    \label{tab:mustc_data}
\end{table}

\begin{table}[h]
    \centering
    \begin{tabular}{lc|rr}
    \toprule
    \textbf{Languages} & \textbf{Code} & \textbf{Hours} &
    \textbf{Samples}
    \\
    \midrule
    English	& En & - & - \\
    French	& Fr & 264 & 207374 \\
    German & De & 184 & 127834 \\
    Spanish	& Es & 113 & 79015\\
    Catalan	& Ca & 136 & 95854 \\
    Italian & It & 44 & 31698 \\
    Russian & Ru & 18 & 12112 \\
    Chinese	& Zh & 10 & 7085 \\
    Arabic	& Ar & 2 & 2283 \\
    Swedish	& Sv & 2 & 2160\\
    Latvian	& Lv & 2 & 2337\\
    Slovenian & Sl & 2 & 1843\\
    Tamil & Ta & 2 & 1358\\
    Japanese & Ja & 1 & 1119\\
    Indonesian & Id & 1 & 1243\\
    Welsh & Cy & 2 & 1241\\
    \bottomrule
    \end{tabular}
    \caption{Statistics of CoVoST-2 X-En involved in this paper.}
    \label{tab:covost2_data}
\end{table}

\begin{table}[h]
    \centering
    \begin{tabular}{c|ccc|c}
    \toprule
    \textbf{Lang.} & \textbf{TED} & \textbf{WMT$^{\dagger}$} & \textbf{CCMatrix$^{\ast}$} &  \textbf{Sum} \\
    \midrule
    \textbf{De} & 0.2M & 4.5M & 43M & 48M \\
    \textbf{Fr} & 0.2M & 39M & 50M & 79M \\
    \textbf{Es} & - & 13M & 61M & 64M \\
    \bottomrule
    \end{tabular}
    \caption{Statistics of monolingual data for MuST-C experiments. $^\dagger:$ De from WMT16~\cite{bojar-EtAl:2016:WMT1}, Fr from WMT14, Es from WMT13, $^\ast:$ randomly sample from CCMatrix~\cite{schwenk2019ccmatrix}.}
    \label{tab:monolingual_data}
\end{table}

\section{Experimental Details}
\label{sec:app:experimental_deatils}
\noindent\textbf{Vocabulary}~
We apply the SentencePiece\footnote{\url{https://github.com/google/sentencepiece}}~\cite{kudo-richardson-2018-sentencepiece} to tokenize the text and discrete units into subwords.  
We add all the discrete units as special symbols to the joint vocabulary. 
The joint subword tokenizer is learned on all the translation sentences and discrete unit sequences in the ST training set.
The vocabulary size is 8000 for both MuST-C and CoVoST-2 experiments. Specifically, 
for MuST-C experiments, since the number of K-means clusters is 500,  the vocabulary is composed of 500 special unit symbols and 7500 text subwords. 
For CoVoST2 X-En experiments, the vocabulary consists of 1000 special unit symbols representing 1000 clusters of mHuBERT,
and 7000 text subwords.

\noindent\textbf{Training details}~
We use Adam optimizer with $\beta_1=0.9, \beta_2=0.98$, and 4k warm-up updates to optimize the parameters in our model. We train the model with a batch size of 5k tokens. The learning rate is 7e-4 and we apply an inverse square root schedule. 
The value of label smoothing is set to 0.1. 
The up-sampling rate $r$ in DUB is set to 32, given the huge volume differences between the BT data and the original data.
For MuST-C experiments, we train U2TT and T2UT models of each translation direction under bilingual settings. 
For CoVoST-2 X-En experiments, we train a multi-lingual X-En model covering 21 translation directions, distinguished by the language tags of the units in different languages.
We implement our models based on
Fairseq\footnote{\url{https://github.com/facebookresearch/fairseq}}~\cite{ott2019fairseq} codebase.
All models are trained on 8 Nvidia Tesla-V100 GPUs and take about 400k steps to converge.
During inference, We save the checkpoint with the best BLEU on the validation set and average the last 10 checkpoints. 
We use beam search with a beam size of 5 for each translation direction.

\noindent\textbf{Training details for Bi-modal BART}~
The training of bi-modal BART follows the recipe of mBART~\cite{liu2020multilingual}. We implemented a mask rate of 0.3, with the replacement of the masked tokens by random tokens at a probability of 0.1. Additionally, the mask length was determined through sampling from a Poisson distribution, with a lambda parameter of 3.5.

\section{Baseline Models}
\label{sec:app:baseline_models}

\noindent\textbf{Existing ST Systems}~
We list the ST systems we compared with on different datasets:
\begin{itemize}[itemsep=1pt, leftmargin=10pt, parsep=0pt, topsep=1pt]
    \item 
    \textbf{MuST-C}~
    In Table~\ref{tab:mustc}, we compare our method with the following: Fairseq ST~\cite{wang-etal-2020-fairseq}, NeurST~\cite{zhao2021neurst}, Espnet ST~\cite{inaguma2020espnet}, E2E-ST-JT~\cite{du2022regularizing}, Speechformer~\cite{papi2021speechformer}, Cascaded ST~\cite{inaguma2020espnet}, MTL~\cite{tang2021improving}, Self-training~\cite{pino2020selfa}, SpeechT5~\cite{ao2022speecht5},  SpeechUT~\cite{zhang2022speechut} and Revisit ST~\cite{zhang2022revisiting}.
    
    \item
    \textbf{CoVoST X-En high resource}~
    In Table~\ref{tab:covost_high}, we compare our method with several baselines from \cite{wang2021covost}, including Transformer-ST, Transformer-ST + ASR pre-train and Cascaded ST, and Revisit ST~\cite{zhang2022revisiting}.
    
    \item
    \textbf{CoVoST X-En low resource}~
    In Table~\ref{tab:covost_low}, we compare our method with several existing ST methods, including Transformer-ST, Transformer-ST + ASR pre-train from~\cite{wang2021covost} and  large-scale multilingual speech or text pre-training methods: XLS-R~\cite{wu2022wav2seq}, Wav2seq~\cite{wu2022wav2seq}, XLS-R+mBART-50~\cite{babu2021xls}, LNA-E,D~\cite{li2021multilingual}.
    Note that XLS-R is pre-trained on 436K hours of speech across 128 languages.
\end{itemize}

\noindent\textbf{Transformer-ST for MuST-C}~
For a fair comparison, we keep the parameters roughly the same size as \method, setting two covolutional layers, a 12-layer Transformer encoder and a 6-layer Transformer decoder, with hidden size $d=768$, $16$ attention heads, and $4096$ FFN hidden states, which makes the model size larger than baselines like Fairseq ST~\cite{wang-etal-2020-fairseq}, NeurST~\cite{zhao2021neurst}, and Espnet ST~\cite{inaguma2020espnet}.

\section{Scalability}
\label{sec:app:ana_model_data_size}


How does model size affect the results of our method? How much improvement does the raw text in the target language bring to our method? To this end, we take MuST-C English-German translation as an example. We set the model size to 73M, 176M and 260M parameters respectively (the specific hyperparameter settings are shown in Table~\ref{tab:model_size}), and introduce extra 1M, 10M, and 48M German sentences.

\begin{figure}[htb]
    \centering
    \includegraphics[width=0.9\linewidth]{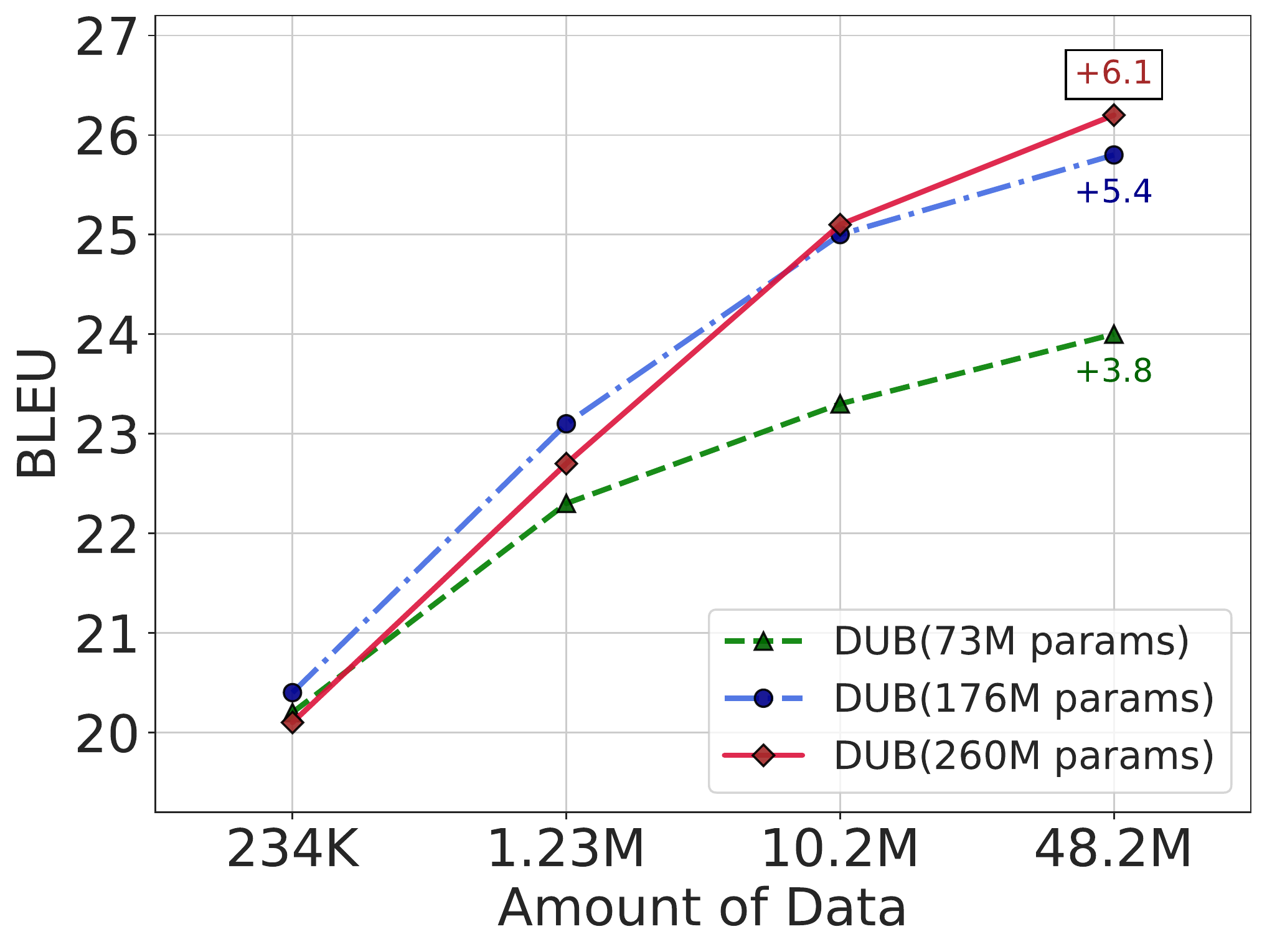}
    \caption{MuST-C En-De \texttt{tst-COM} BLEU scores for different amounts of target-language text data are introduced, under different model sizes.}
    \label{fig:data_model_size}
\end{figure}

Figure~\ref{fig:data_model_size} shows the BLEU scores of different sizes of models, with different amounts of monolingual back-translation data added.
In general, regardless of the model size, introducing \textbf{more text brings better performance}. When we introduce a large amount of back-translated data, \textbf{the larger model gets significantly better performance}. We find that when no or less back-translated data is introduced, the performance of the large model is instead not optimal. This is because the large model is prone to overfitting when the original training data is small, but as the monolingual data is gradually introduced, the advantage of the large model becomes obvious, without replying to the transcription, introducing 48M back-translated pairs, the model with 260M parameters can boost up to 6.1 BLEU on En-De.

\begin{table}[htb]
    \small
    \centering
    \begin{tabular}{lc|ccc}
        \toprule
         & \textbf{Model} & \textbf{Encoder} & \textbf{Decoder} & \textbf{Hidden} \\
         & \textbf{Params} & \textbf{Layers} & \textbf{Layers} & \textbf{Dim}\\
        \midrule
        \textbf{\circled{1}}\textsc{Small} & \textbf{73M} & 6 & 6 & 512 \\\hline
        \textbf{\circled{2}}\textsc{Base} & \textbf{176M} & 12 & 6 & 768 \\\hline
        \textbf{\circled{3}}\textsc{Large} & \textbf{260M} & 12 & 6 & 1024 \\
        \bottomrule
    \end{tabular}
    \caption{Hyper-parameter settings for the models in Figure~\ref{fig:data_model_size}.}
    \label{tab:model_size}
\end{table}

\section{Comparison With Cascaded System}
\label{sec:app:comp_with_cascaded}
It could be argued that our model employs a cascaded architecture, comprising a unit extractor and a unit-to-text translation model.
The traditional cascade ST system (ASR+MT) can also be enhanced through applying back-translation to improve its MT model. 

In Table \ref{tab:comaprision_with_cascaded}, we compare the performance of DUB with the BT-enhanced cascaded ST system both utilizing 10M unpaired text.
By comparison, we can find that the BLEU score of the U2TT model is inferior to that of the cascaded system when utilizing 10 million unpaired text samples. This discrepancy can likely be attributed to the higher baseline performance of the cascaded system. Additionally, DUB demonstrates a superior relative improvement in BLEU score compared to the cascaded system. Moreover, the discrete unit extractor is obtained through unsupervised training on unlabeled speech, which requires no transcriptions compared with the ASR system trained on speech-transcription pairs.

\begin{table}[]
    \centering
    \begin{tabular}{l|c|ccc}
        \toprule
        \textbf{Method} & \textbf{Extra Text} & \textbf{BLEU} & \textbf{$\Delta$BLEU}\\
        \midrule
        U2TT & - & 20.4 & -\\
        \quad w/ DUB & \checkmark & 25.0 & 4.6\\
        \midrule
        Cascaded ST & - & 23.1 & - \\
        \quad w/ MT-BT & \checkmark & 26.0 & 2.9 \\
        \bottomrule
        
    \end{tabular}
    \caption{MuST-C En-De \texttt{tst-COM} BLEU for U2TT and cascaded system. MT-BT refers to enhancing MT model of cascaded ST through back-translation. Cascaded ST is trained by ourselves on MuST-C En-De \texttt{train} set. The same extra text corpus with 10 million German sentences is used for both methods.}
    \label{tab:comaprision_with_cascaded}
\end{table}

\section{Cases of Text-to-Speech Translation}

\label{sec:app:cases_tts_translation}

In Table \ref{tab:cases_tts_translation}, we show two cases of German-English text-to-speech translation on MuST-C En-DE \textsc{tst-COM} set.
In CASE 1, our text-to-speech translation system generates speech with the same content and a similar spectrogram as reference speech.
In CASE 2, the synthetic speech deviated slightly from the reference speech, but the translation is correct --- ``release'' has the same meaning as ``shoveling out'' and ``all the time '' just means ``all along''.
The samples of generated audio are included in \url{https://anonymous.4open.science/r/DUB/ttss_samples}.

\clearpage
\begin{table*}[]
    \small
    \centering
    \resizebox{\textwidth}{!}{
    \begin{tabular}{l|p{13.cm}}
        \toprule
        & \textbf{German:} Einzige Land der Welt. \\
         
        & \textbf{Ref:} the only country in the world.\\
        &  
        \begin{minipage}[t]{\textwidth}
            \vspace{0pt}
            {\includegraphics[width=0.9\textwidth]{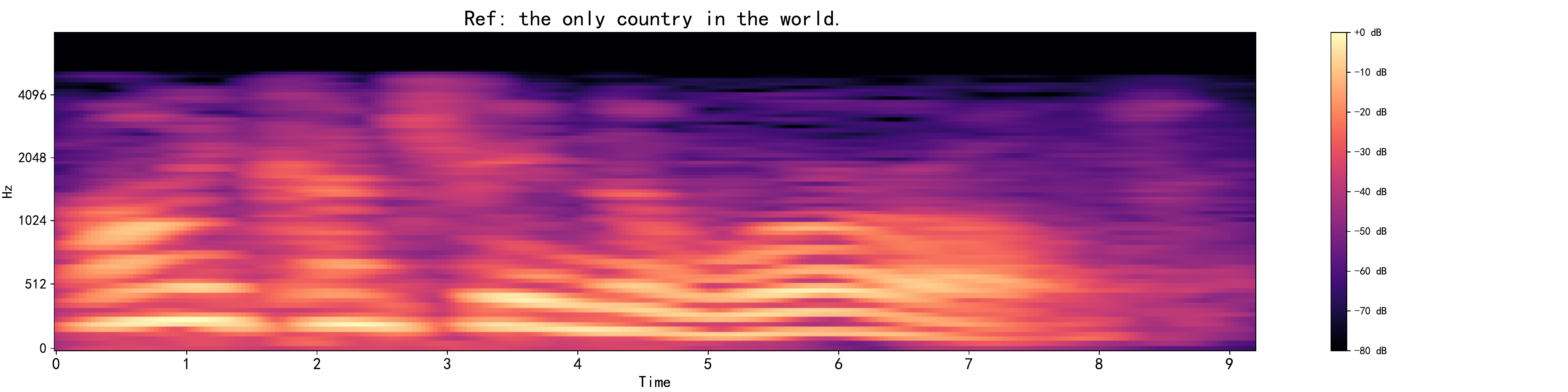}}
        \end{minipage} \\
        
        \textbf{CASE 1} & \textbf{Hyp:} the only country in the world.\\
        &
         \begin{minipage}[t]{\textwidth}
            \vspace{0pt}
            {\includegraphics[width=0.9\textwidth]{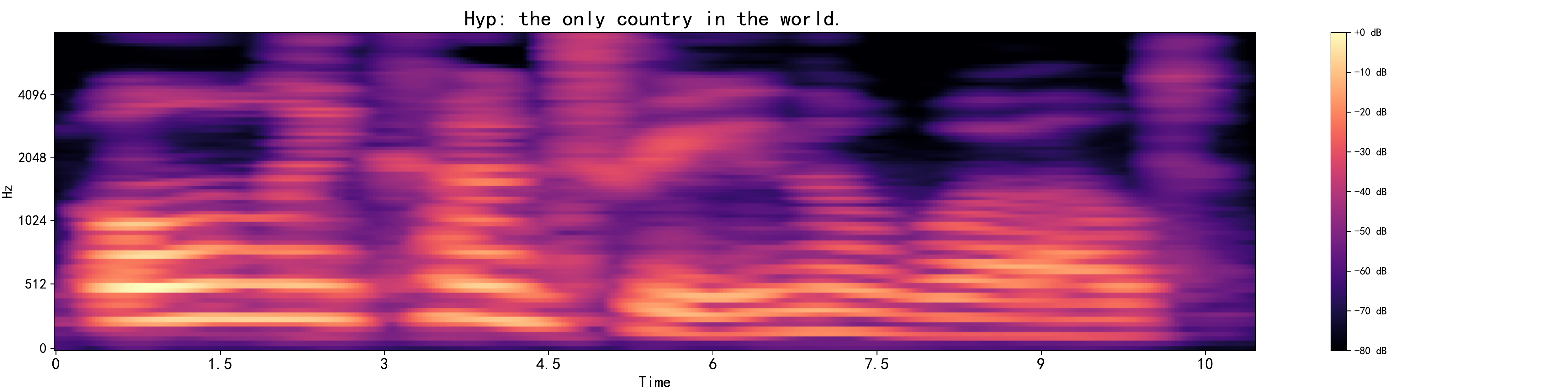}}
        \end{minipage} \\

        \midrule
         & \textbf{German:} Und sicher, hätten wir diese diese Hydranten die ganze Zeit freischaufeln können, und viele Menschen tun das. \\
         
          & \textbf{Ref:} And certainly, we could have been shoveling out these fire hydrants all along, and many people do.\\
        &
        \begin{minipage}[t]{\textwidth}
            \vspace{0pt}
            {\includegraphics[width=0.9\textwidth]{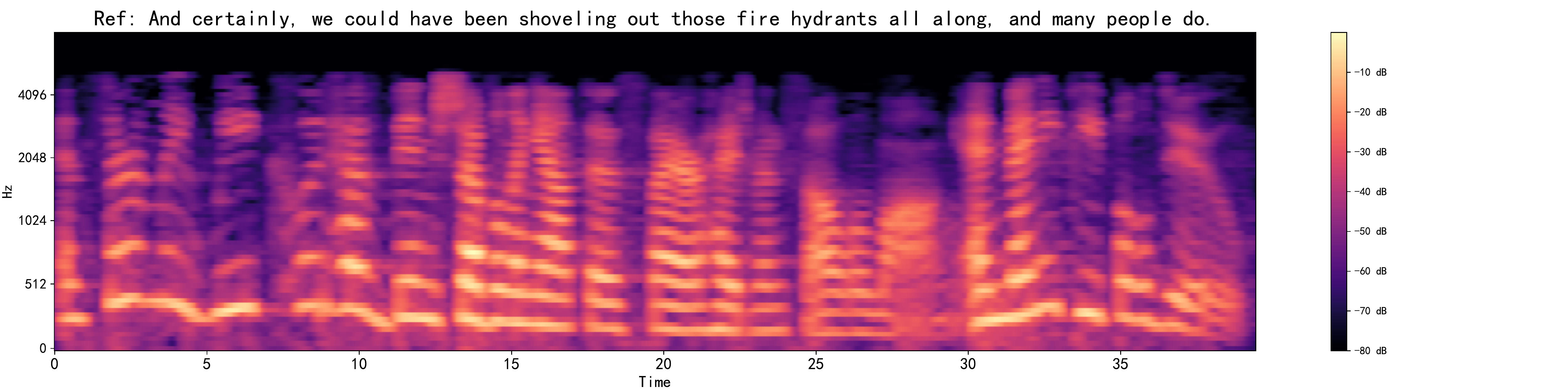}}
        \end{minipage} \\
        
        \textbf{CASE 2} & \textbf{Hyp:} And certainly we could have released those hydrants all the time and many people do that.\\
        &
         \begin{minipage}[t]{\textwidth}
            \vspace{0pt}
            {\includegraphics[width=0.9\textwidth]{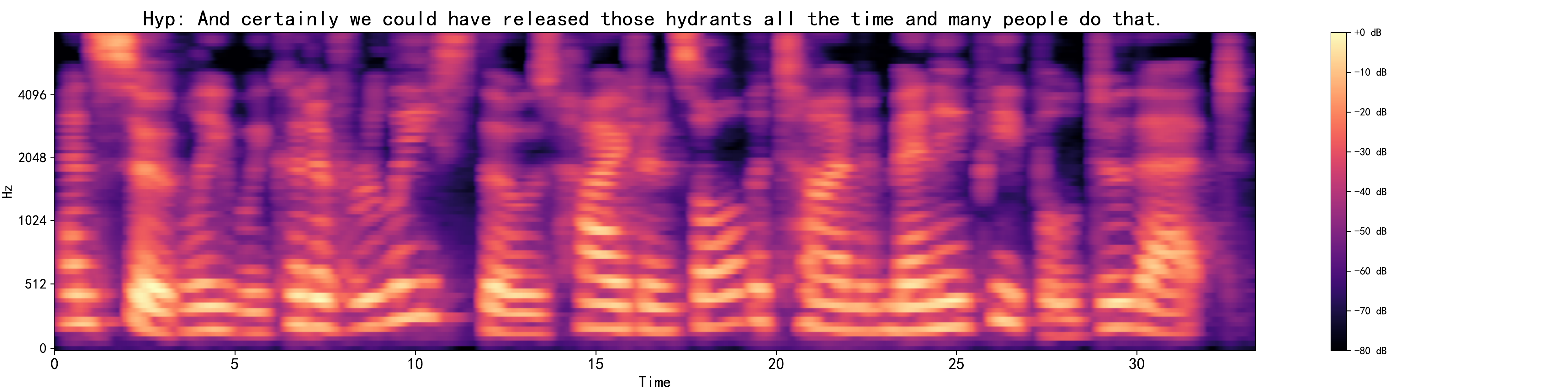}}
        \end{minipage} \\
        
        \bottomrule
    \end{tabular}
    
    }
    \caption{Cases of De-En text-to-speech translation on MuST-C En-De \texttt{tst-COM} set. 
    }
    
    \label{tab:cases_tts_translation}
\end{table*}

\end{document}